\documentclass[fleqn,twoside]{article}
\usepackage[headings]{espcrc2}

\readRCS
$Id: espcrc2.tex,v 1.2 2004/02/24 11:22:11 spepping Exp $
\usepackage{graphicx, balance}
\usepackage{epstopdf}
\usepackage{subfigure}
\usepackage[linesnumbered, ruled, vlined]{algorithm2e}
\usepackage{url}
\usepackage[figuresright]{rotating}

\newcommand{\AmS}{{\protect\the\textfont2
  A\kern-.1667em\lower.5ex\hbox{M}\kern-.125emS}}

\title{\textbf{Reinforcement Learning approach for Real Time Strategy Games Battle city and S3}}

\author{Harshit Sethy\address{CTO of Gymtrekker Fitness Private Limited,Mumbai, India, Email: hsethy1@gmail.com\\},
Amit Patel\address[DCSE]{Assistant Professor, Department of Computer Science and Engineering, RGUKT IIIT Nuzvid, Krishna-521202 India, Email: amtptl93@gmail.com}}

\runtitle{Reinforcement Learning approach for Real Time Strategy Games like Battle city and S3}
\begin{document}
\begin{abstract}
In this paper we proposed reinforcement learning algorithms with the generalized reward function. In our proposed method we use Q-learning and SARSA algorithms with generalised reward function to train the reinforcement learning agent. We evaluated the performance of our proposed algorithms on two real-time strategy games called BattleCity and S3. There are two main advantages of having such an approach as compared to other works in RTS. (1) We can ignore the concept of a simulator which is often game specific and is usually hard coded in any type of RTS games (2) our system can learn from interaction with any opponents and quickly change the strategy according to the opponents and do not need any human traces as used in previous works.

\end{abstract}

\maketitle

\section{INTRODUCTION}
\label{section:Introduction}
Existence of a good artificial intelligence(AI) technique in the background of a game is one of the major factor for the fun and re-play ability in commercial computer games. Although AI has been applied successfully in several games such as chess, backgammon or checkers when it comes to real-time games the pre-defined scripts which is usually used to simulate the artificial intelligence in chess, backgammon etc~\cite{Ponsen}. does not seem to work. This is because in real-time games decisions has to be made in real-time as well as the search space is huge and as such they do not contain any true AI for learning~\cite{Genter}. Traditional planning approaches are difficult in case of RTS games because they have various factors like huge decision spaces, adversarial domains, partially-observable, non-deterministic  and real-time, (real time means while deciding the best actions, the game continues running and states change simultaneously).

\subsection{Real Time Strategy Games}
\begin{figure}[h]
\centering
\includegraphics[width=6cm]{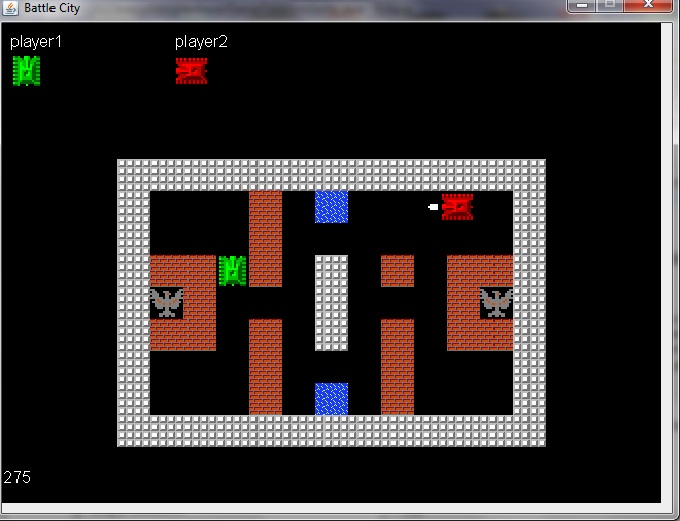}\\
(a)\\
\includegraphics[width=6cm]{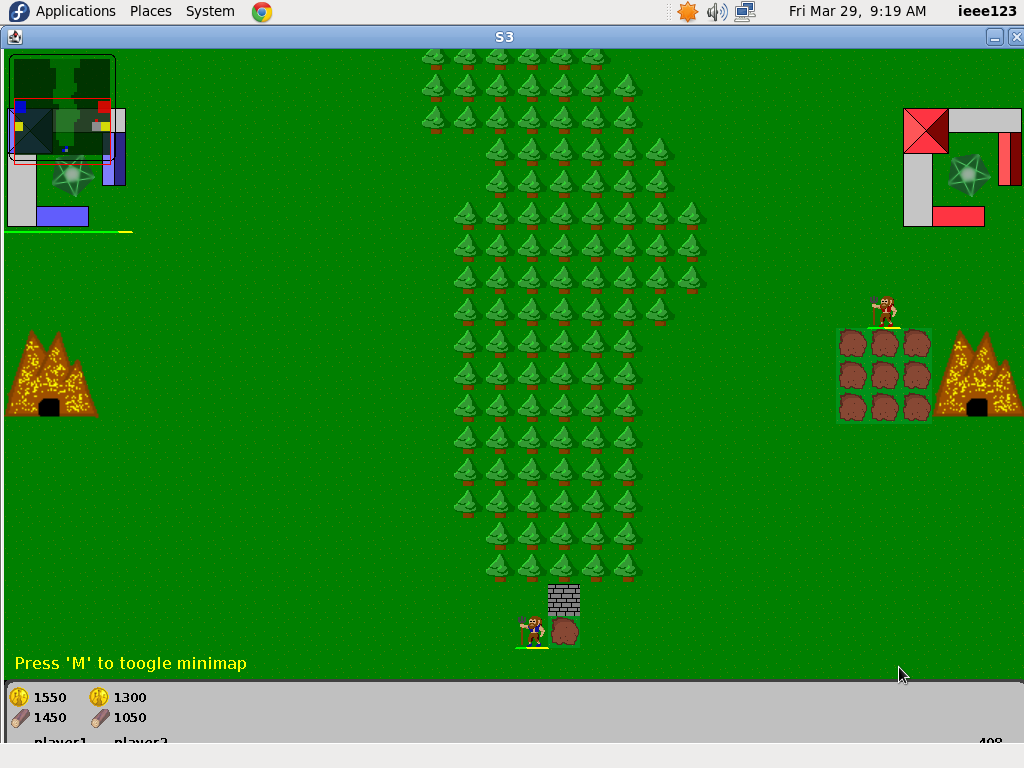}\\
(b)
\caption{(a)Snapshot of a BattleCity Game (b)Snapshot of an S3 Game} 
\label{BattleCity_S3}
\end{figure}

Today game developing companies have started showing more interest in RTS games. Unlike turn based strategy games, where one has the ability to take ones own time, in real time strategy games, all movement, construction, combat etc., are all occurring in real time. In a typical RTS game, the screen contains a \texttt{map} area which consists of the game world with buildings, units and terrain. There are usually several players in an RTS game. Other than the players there are various game entities called \emph{participants, units} and \emph{structures}. These are under the control of the players and the players need to save their assets and/or destroy assets of the  opponent players by making use of their control over the entities. We are using 2 RTS games (1) BattleCity and (2) S3 game for our evaluation. A snapshot of two RTS games called \texttt{BattleCity} and \texttt{S3} are given in Figure \ref{BattleCity_S3}.

\subsection{BattleCity Game}
\texttt{BattleCity} is a multidirectional shooter video game, which can be played using two basic actions Move and Fire. The player, controlling a tank, must destroy enemy tanks or enemy base and also protect its own base. Player can move tank in four directions (left, right, up and down) and fire bullets in whichever direction the tank last moved, while bases are static. There are three types of obstacle. (1) \textit{Brick wall} – tank can destroy it by firing this type wall. (2) \textit{Marble wall} – tank can’t destroy it by firing. (3) \textit{Water bodies} – tank can fire through it. Tank can’t pass through any of above obstacle. Only brick wall can be destroyed by tank so after destroying tank can pass through it.
\\

%\begin{figure*}[ht]
%\centering
%\includegraphics[width=8.5cm]{bc_game.jpg}
%\caption{Snapshot of a BattleCity Game} 
%\label{BattleCity}
%\end{figure*}

\subsection{S3 Game}
\texttt{S3} is a real-time strategy game where each player’s goal is to remain alive after destroying the rest of the players. Four basic actions in this game are \textit{Harvest}: i.e., to gather resources (gold and wood), \textit{Build}: to build buildings (Barrack, Blacksmith, Tower etc) ,\textit{Train}: to produce troops (archers, footmen, catapults, knights), \textit{Attack}: for attacking enemy. 

%\begin{figure*}[ht]
%\centering
%\includegraphics[width=8.5cm]{Screenshot.png}
%\caption{Snapshot of an S3 Game} 
%\label{S3}
%\end{figure*}

\paragraph{}
This paper is structured as follows. Apart from introduction, there are five more sections.In section 2 highlights the review of related works. In section 3 we discuss about reinforcement learning techniques in real-time-strategy games and outline the various learning algorithms used in reinforcement learning. In section 4 we outline implementation details related to the proposed reinforcement learning algorithms with the generalized reward function for two real-time-strategy games (1) BattleCity and (2) S3 game. Section 5 discusses about the experimental result related to our proposed work for BattleCity and S3. We conclude with section 6.

\section{Related Work}
\label{section:related work}

One of the major works using \emph{Online case-based planning}~\cite{Janet} techniques %is the first major work using in 
for Real Time Strategy Games was published in~\cite{Neha}. On-line case-based planning revises \texttt{case based planning} for strategic real-time domains involving on-line planning. 

In~\cite{Onta} a case-based planning system called Darmok2 is introduced that can play RTS games. They introduced a set of algorithms that can be used to learn plans, represented as \texttt{petri-nets}, from one or more human demonstrations.
Another work by the same authors which uses Darmok2 but addresses the issues of plan acquisition, on-line plan execution, interleaved planning and execution and on-line plan adaptation is~\cite{Santi}. 

In ~\cite{AshwinRam} the authors summarize their work in exploring the use of the first order inductive learning (FOIL) algorithm for learning rules which can be used to represent opponent strategies.

In ~\cite{Pranay} the authors improve  Darmok2 using information related to sensors of the game. We refer to that work as PR-Model in this paper. PR-model is capable of learning how to play RTS games by observing human demonstrations. Using human traces PR-model makes plans to play games. Prioritize the plan according to the feedback of the game and feedbacks are decided using some rule which depends on the sensors of the game. 

Drawbacks of all case based learning~\cite{Gomez} approaches as mentioned above are (1) It requires expert demonstrations for making plans (2) after training is done, no further learning takes place (3) to cover large state spaces it would require large number of rules in the plan base (4) no exploration for optimal solution. Only follows human traces.. Stefan Wender~\cite{Stefan} uses Reinforcement Learning for City Site Selection in the Turn-Based Strategy Game Civilization IV. Civilization IV is the strategy game it is a turn-based game while Battle City is Real time game.

Stefan Wender~\cite{Stefan} uses Reinforcement Learning for City Site Selection in the Turn-Based Strategy Game Civilization IV. Civilization IV is the strategy game similar to S3 but it is a turn-based game while S3 is Real time multi agent game.

 In this paper we aim to do away with the hard coded simulator and propose a learning approach based on \texttt{Reinforcement Learning}~\cite{RSSuttan}(RL) wherein sensor information from the current game-state is used to select the best action.
Reinforcement learning is used because of its advantages over previous strategies. Specifically (1) RL cuts out the need to manually specify rules. RL agents learn simply by playing the game against other human players or even other RL agents (2) for large state spaces, RL can be combined with a function approximator such as a neural network, to approximate the evaluation function (3) RL agent always explores for optimal solution to reach the goal (4) RL has been applied widely to many other fields, such as robotics, board games ,turn based games and single agent games with great results, but hardly ever on RTS multi-agent games.

\section{Reinforcement Learning}

\textit{Reinforcement Learning}~\cite{RSSuttan} is the field of \textit{Machine Learning} which deals with what to do, how to map situations to actions so as to maximize a numerical reward signal.The learner does not know which actions to take, as in most forms of machine learning, but instead must discover which actions gives the most reward by applying them. In the most interesting and challenging cases, actions may affect not only the immediate reward but also the next situation and, through that, all subsequent rewards.

	With comparing reinforcement learning~\cite{Marthi} to RTS game environment an AI player learns by interacting with the environment and observing the feed-backs of these interactions. This is same as the fundamental way in which humans (and animals) learn. As a human, we can perform actions and observe the results of these actions on the environment. The same way RL-agent interacts with the environment and observes the result and assign the reward or penalty to state or state-action pair according to the desirability of the resultant state.

\begin{figure}[h]
\centering
\includegraphics[scale=.28]{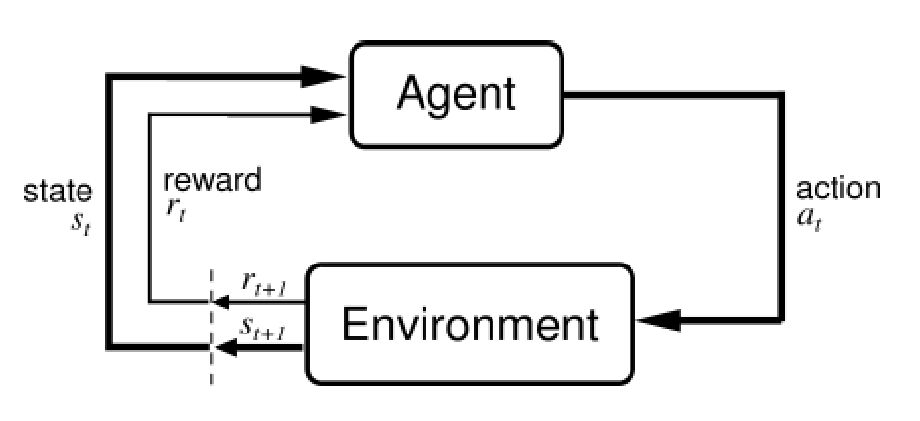} \\
(a)\\
\includegraphics[width=2.8in]{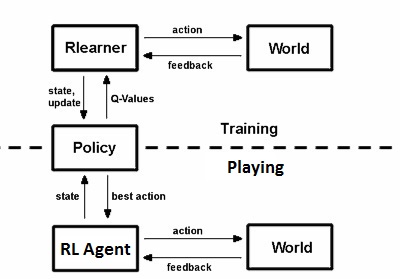} \\
(b)
\caption{(a)Reinforcement Learning (b) Architecture for the Reinforcement Learning}
\label{Reinforcement Learning}
\end{figure}

%------------------------------------

\subsection{Reinforcement Learning Architecture}
\textit{RL Architecture} has two main characteristics; one is learning and the other is playing with the learnt experiences. 
Initially RLearner has no Knowledge about the game. So it does random actions and observe the resultant state using some sensor information of the game and give feedback (in the form of reward which is further used to calculate the Q-Values for the state-action pairs or Q-Table) of that action to the previous state according to the desirability of the current state. Q-Values of the state-action pairs are known as Q-Table which define a policy. After every action policy updates Q-Values for the state action pairs (Q-Table) this policy is used to predict the best action while playing the game. RL agent learns while playing so it again gives feedback and the whole process it going on till the end of the game.  

%--------------------------------------------------------------------

%\newpage
%\subsection{Basic components of RL \& relation with BattleCity and S3}
\subsection{Basic components of RL}
Reinforcement learning contains five basic components which are as listed below.
\begin{enumerate}
\item a set of environment states S 
\item a set of actions A 
\item rules of transitioning between states
\item rules that determine the scalar immediate reward of a transition (Reward Functions)
\item rules that describe what the agent observes (Value Functions)
\end{enumerate}

\subsubsection{Reward  Function} The scalar value which represents the degree to which a state or action is desirable is known as \textit{reward}. 
This scalar reward is assigned to the action for the particular transition and the resultant state of the game. If the resultant state is desirable and safe then positive scalar value as reward will be assigned to that action otherwise if state is not safe or undesirable then some negative scalar value as negative reward will be assigned to that action. 
We are using 2 types of Reward function (1) \emph{Conditional Reward function} (2) \emph{Generalised Reward function}.

\subsubsection{Value Function}
\textit{Value Functions} are used for mapping from states or from state-action pairs to real numbers, where the value of a state represents the long-term reward achieved starting from that state (or state-action), and executing a particular policy. 
It estimates how good a particular action will be in a given state, or what the return for that action is expected to be. There are two type of value functions. 
\begin{enumerate}
\item $V^{\pi}(s)$ is the value of a state '$s$' under policy $\pi$. The expected return when starting in s and following $\pi$ thereafter. 
\item $Q^{\pi}(s,a)$ is the value of taking action '$a$' in state '$s$' under a policy $\pi$. The expected return when starting from s taking the action a and thereafter following policy $\pi$.  
\end{enumerate}

%\subsubsection{How these value functions are calculating the values}
There are two methods to define these value functions:
\begin{enumerate}
\item \emph{Monte Carlo~\cite{RSSuttan} Method}: In this method the agent would need to wait until the final reward was received before any state-action pair values can be updated. Once the final reward is received, the path taken to reach the final state would need to be traced back and each value updated.
\begin{equation}
V(s_{t}) \leftarrow V(s_{t})+\alpha[R_{t}-V(s_{t})]
\end{equation}
\hspace{6mm} where $s_{t}$ is the state visited at time t, $R_{t}$ is the reward after time t and $\alpha$ is a constant parameter.

\item \emph{Temporal Difference~\cite{RSSuttan} Method}: It is used to estimate the value functions after each step. An estimate of the final reward is calculated at each state and the state-action value updated for every step of the way. This reflects a more realistic assignment of rewards to actions compared to MC, which updates all actions at the end directly. TD Learning is  nothing but the combination of dynamic programming with the Monte Carlo method. The formula related to TD learning is given as 
\begin{equation}
V(s_{t}) \leftarrow V(s_{t})+\alpha[r_{t+1}+\gamma V(s_{t+1})-V(s_{t})]
\end{equation}
\hspace{6mm} where $r_{t+1}$ is the observed reward at time t+1.\\\\
\end{enumerate}

\subsection{Sensor representation for S3 and BattleCity Game}
\begin{figure}[h]
\centering
\mbox{\subfigure{\includegraphics[width=1.5in, height=1.4in]{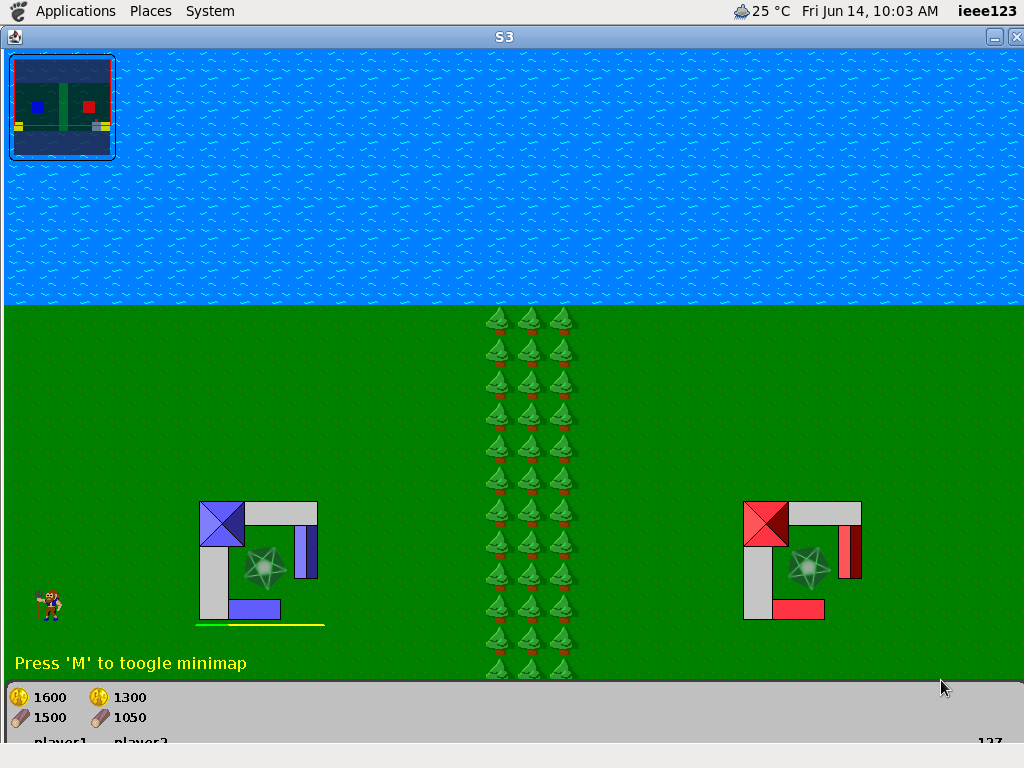}}
\subfigure{\includegraphics[width=1.1in]{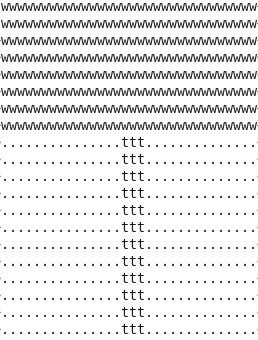} }}
%\caption{Snapshot of S3, BattleCity Games and there current 2D maps} \label{snapshot of BattleCity and 2D map}
\end{figure}
\begin{figure}[h]
\centering
\includegraphics[width=2.7in]{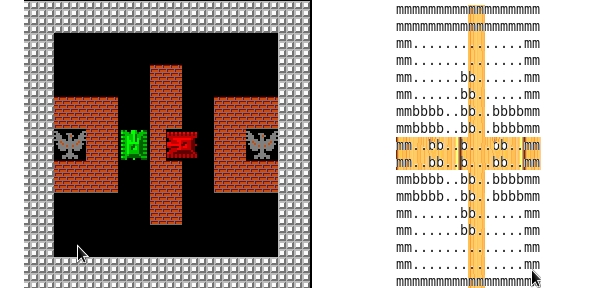}
\caption{Snapshot of S3, BattleCity Games and there current 2D maps} 
\label{snapshot of BattleCity and 2D map}
\end{figure}

We are using two types of sensor information for assigning reward in battle city game which are explained as follows;
\begin{enumerate}
\item \textit{EnemyInline}: If \emph{enemy position} is directly in line with player without any block or wall then sensor is represented by number 2. If there is a wall or block between enemy and player then sensor is represented by number 1. If enemy position is not in line with player then sensor is 0.
\item \textit{EnemyBaseInline}: This sensor information is represented in the same way as above but instead of taking into consideration position of enemy, position of enemy-base is taken into account. If \emph{enemy-base position} is directly in line with player without any block or wall then sensor is represented by number 2. If there is a wall or block between enemy-base and player then sensor is represented by number 1. If enemy-base position is not in line with player then sensor is 0. 
\end{enumerate}

Sensor information for S3 game
\begin{enumerate}
\item Get the current map and store it in a two dimensional array.
\item Gold and Wood sensors are retrieved from current game-state.
\item Number of peasant and footmen entities for both enemies and player are retrieved from entities state.
\item Update two dimensional array with static entities like goldmine position with 'g', and buildings with 'b'.  
\end{enumerate}

So far we have outlined our method of obtaining sensor information related to two real-time strategy games, BattleCity and S3. 
%In the next section we show how the reward is calculated based on the acquired sensor information.

\subsection{Action Selection Policies}
We have the following action selections policies which can be used to select desired action according to the behavior of that particular policy
\begin{enumerate}
\item \textbf{$\epsilon-$greedy} : Most of the time the action with the highest estimated reward is chosen, called the greediest action. But, with a small probability $\epsilon$, an action is selected at random to ensure optimal actions are discovered.
\item \textbf{$\epsilon-$soft} : Very similar to $\epsilon-$greedy. The best action is selected with probability $1 - \epsilon$ and the rest of the time a random action is chosen uniformly. 
\item \textbf{softmax} : One drawback of the above methods is that they select random actions with some probability. So there is a case when the worst possible action is selected as the second best. Softmax remedies this by assigning a rank or weight to each of the actions, according to their action-value estimate. So the worst actions are unlikely to be chosen. 
\end{enumerate}

\subsection{Steps while learning}
\begin{enumerate}
\item The Rlearner observes an input Game state.
\item The Rlearner then creates a new policy based on the dimensions of the world.
\item Set the parameters ($\alpha,\gamma,\epsilon$ and number of episodes) for the Rlearner and start learning.
\item Start running epochs. You can optionally run each epoch individually.
\end{enumerate}
One epoch contains following steps.
\begin{enumerate}
\item An action is determined by a decision making function (e.g. $\epsilon-$greedy).
\item The action is performed.
\item The Rlearner receives a scalar reward or reinforcement from the environment according to reward function.
\item Information about the reward given for that state / action pair is recorded.
\item Update the Q-values in Q-table According to Learning Algorithm(e.g. Q-learning or SARSA).
\end{enumerate}

\section{Proposed learning algorithm}
In this section we outline our proposed learning algorithms which we integrated into the two RTS games Battlecity and S3. We also provide the implementation details related to selection of parameters and reward functions.

\subsection{Parameters}
This section contains the information regarding the reward algorithms and its parameters which we use for the two game BattleCity and S3. 
 
\begin{itemize}
\item \textbf{Learning Rate $\alpha$ :} The learning rate $0 < \alpha < 1$ determines what fraction of the old estimate will be updated with the new estimate. $\alpha = 0$ will stop the RL-agent from learning anything while $\alpha = 1$ will completely change the previous values with the new one.
\item \textbf{Discount Factor $\gamma$ :} The discount factor $0 < \gamma < 1$ determines what fraction of the upcoming reward values will be considered for evaluation. For $\gamma = 0$ all the upcoming rewards are ignored. For $\gamma = 1$ means the RL-Agent will consider the current and upcoming rewards as equal weightage. 
\item \textbf{Exploration Rate $\epsilon$ :} In action selection policies 
%(Section 3.7) 
there is one policy called as $\epsilon$ greedy method which uses the exploration rate $0 < \epsilon < 1$ for determining the ratio between the exploration and exploitation. We are using $\epsilon$ greedy method for selecting the best action and to maintain the balance between exploration and exploitation.
\end{itemize}

\subsection{Reward function for BattleCity}
\textbf{Algorithm \ref{calcReward_algo}}: Reward function is for calculating reward after performing action on current state. According to the result of the action reward or penalty are assigned. In steps 1 to 9 get the positions (x-y co-ordinates) of the player, enemy and enemy base on the map. In steps 10 to 16 if game is over and winner is the RL-Agent (player) then add the reward to the total reward (newReward) else deduct penalty from the total reward. In steps 17 to 18 if enemy is in line with the RL-Agent deduct penalty from total reward so it always tries not to be in line with enemy. In steps 19 to 21 if enemy base is in line with the RL-Agent  then calculate the distance between the enemy base and RL-Agent and deduct from 2 times of reward and add to total reward. So it pushes the RL-Agent to come closer to the enemy base. Steps 22 to 24 gives the generalized reward function which makes the RL-Agent quickly attack the enemy base and prevent attack by the enemy.
 
\%%%%%%%%%%%%%%%%% Algorithm %%%%%%%%%%%%%%%%%%%
\begin{algorithm*}[ht]
\begin{small}
\KwIn{$state$ :- contains positions of entities, reward, penalty \\
$sensorsList$ :- contains sensors of game domain.\\
$gameState$ :- contains state of game is running or not\\}
\KwOut{Reward}
\BlankLine
 $Player_x$ = null, $Player_y$ = null, $Enemy_x$ = null, $Enemy_y$ = null \;
 $EnemyBase_x$ = null, $EnemyBase_y$ = null, $winner$ = null \;
 $newReward$ = 0, $distance$ = 0 \;
 $Player_x$ = \texttt{getPositionx($state$,player)} \; 
 $Player_y$ = \texttt{getPositiony($state$,player)} \;
 $Enemy_x$ = \texttt{getPositionx($state$,enemy)}  \;
 $Enemy_y$ = \texttt{getPositiony($state$,enemy)} \;
 $EnemyBase_x$ = \texttt{getPositionx($state$,enemybase)} \;
 $EnemyBase_y$ = \texttt{getPositiony($state$,enemybase)} \;
 \If {$gameState$ == \texttt{"end"}}{
	 $winner$ = \texttt{getWinner()} \;             	
	\If {$winner$ == \texttt{"player"}}{
		$newReward$ = $newReward$ + $reward$ \; }           	
	\Else{
		$newReward$ = $newReward$ - $penalty$ \;  }         			
 }
\Else {
	\If {sensorList[EnemyInline]==2}{
		$newReward$ = $newReward$ - $penalty$ \;
	}
	\If {sensorList[EnemyBaseInline]==2} {
		$distance$ = $\sqrt[2]{(EnemyBase_x - Player_x)^{2} + (EnemyBase_y - Player_y)^{2}}$ \;
		$newReward$ = $newReward$ + $2 \times reward$ - $distance$ \;
	}
	$newReward$ = $newReward$ - $4 \times distance$ \;
	$distance$ = $\sqrt[2]{(Enemy_x - Player_x)^{2} + (Enemy_y - Player_y)^{2}}$ \;
	$newReward$ = $newReward$ + $4 \times distance$ \;
 }	          
\Return $newReward$ \;
\caption{calcReward for BattleCity }
\label{calcReward_algo}
\end{small}
\end{algorithm*}
%%%%%%%%%%%%%%%%%%%%%%%%%%%%%%%%%%%%%%%%%%%%%%%%%%%%%%%%%%

\subsection{Reward function for S3}
\textbf{Algorithm \ref{calcReward_S3_algo}}: In step 1 to 6 get the sensors related to total gold, total wood and size of troops of the player and enemy. In steps 7 to 11 if game is over and winner is the RL-Agent (player) then add the reward to the total reward (newReward) else deduct penalty from the total reward. In steps 12 to 14 and 17 to 18 if gold and wood for player is greater than enemy than add reward to the total reward otherwise deduct penalty from total reward so it always tries to increase the gold and wood with compare to enemy. In steps 21 to 22 if Player troop is bigger than the Enemy troop then add the twice of reward to the total reward (newReward) else deduct twice of penalty from the total reward. So it pushes the RL-Agent to Attack or build the army to increase the size of troop as compared to the enemy. In step 25 Return the total reward.

\begin{algorithm*}[ht]
\begin{small}
\KwIn{$state$ :- contains positions of entities, reward, penalty\\
\textbf{Global access to: } $sensorsList$ :- contains sensors of game domain\\
$gameState$ :- contains state of game is running or not\\}
\KwOut{Reward}
\BlankLine

$Player_g$ = 0, $Player_w$ = 0, $Enemy_g$ = 0, $Enemy_w$ = 0
$EnemyTroopLength$ = 0, $PlayerTroopLength$ = 0, $winner$ = null
$newReward$ = 0
$Player_g$ = \texttt{player.getGold()} \;  
$Player_w$ = \texttt{player.getWood()} \;
$Enemy_g$ = \texttt{enemy.getGold()}  \;
$Enemy_w$ = \texttt{enemy.getWood()} \;
$EnemyTroopLength$ = \texttt{enemyTroop.size()} \;
$PlayerTroopLength$ = \texttt{playerTroop.size()} \;
	\If {$gameState$ == \texttt{"end"}}{
		     $winner$ = \texttt{getWinner()}             	
		\If {$winner$ == \texttt{"player"}}{
			     $newReward$ = $newReward$ + $reward$}            	
		\Else{
			     $newReward$ = $newReward$ - $penalty$   }
		}
	\Else {
		\If {$Player_g > Enemy_g$}{
			     $newReward$ = $newReward$ + $reward$     }       	
		\Else {
			     $newReward$ = $newReward$ - $penalty$   }
				
		\If {$Player_w > Enemy_w$}
			{    $newReward$ = $newReward$ + $reward$}            	
		\Else
			   { $newReward$ = $newReward$ - $penalty$}
			   
		\If {$PlayerTroopLength > EnemyTroopLength$}{
			      $newReward$ = $newReward$ + 2*$reward$ }           	
		\Else {
			     $newReward$ = $newReward$ - 2*$penalty$ }
			
	}	          
\Return $newReward$
\caption{ calcReward for S3}
\label{calcReward_S3_algo}
\end{small}
\end{algorithm*}

\section{Experimental Results}
In the previous section we have discussed how we successfully applied reinforcement learning in two real-time strategy games called BattleCity and S3. In this section we outline the experimental results related to reinforcement learning in BattleCity and S3. 

\subsection{BattleCity:}
We evaluated the performance of RL-Agent with the help of various maps (e.g. \emph{Bridge-26x18, Bridge-metal-26x18, Bridges-34x26}) as well as with two types of opponents called \emph{AI-Random} and \emph{AI-Follower} in each map. We observed that the Reinforcement Learning Agent won more than 90\% games when played against both opponents( AI-Random and AI-Follower) in simple maps and about 80\% to 90\% when played against AI-Random in complex maps and 60\% to 80\% when played against AI-Follower in complex maps. Statistics about the performance of the SARSA\cite{RSSuttan}, Q-Learning\cite{RSSuttan} and Darmok2 in the various maps are represented below in the form of graphs. We observed that performance of RL-Agent under SARSA Learning algorithm is better than other techniques and also RL-agent trained by SARSA algorithm takes less time to win the game.

%%%%%%%%%%%%%%%%%%%%%
We performed our evaluation for BattleCity game against two opponents AI-Random and AI-Follower with three different maps. AI-Random is the built-in AI which selects random action always and AI-Follower is tough to compete because it always follows the opponent and fires at it. It is clear from the experimental results that reinforcement learning agent with the SARSA~\cite{RSSuttan} algorithm performs better than other techniques like Q-Learning~\cite{RSSuttan} and online case based learning based on Darmok2~\cite{Santiago}. Statistics related to performance are given  below in the form of graphs. Statistics are represented using two types of graphs. One is time (in milliseconds) taken to win the game versus episodes. X-axis represents the number of episode and Y-axis represents the time in milliseconds. The other is number of games won versus episode. Here also X-axis represents the number of episodes and Y-axis represents the total number of games won till that episode.
%%%%%%%%%%%%%%%%%%%%%%%%%%%%

\subsubsection{Map: Bridge-26x18}
\begin{figure}[h]
\centering \includegraphics[width=2.6in]{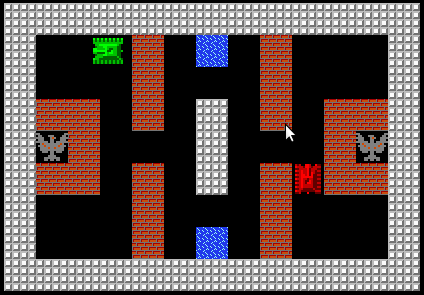}
\caption{Map:Bridge-26x18} \label{Map:Bridge-26x18}
\end{figure}

This map size is 26x18 (refer Figure \ref{Map:Bridge-26x18}) so total state space for this map is total combination of the $x-y$ co-ordinates of the player and enemy which is $26^{2}$x$18^{2}$. This map has a marble wall in between which the tank cannot destroy by firing. So this is an advantage for the tank to hide from opponents and attack when opponents enters their side.

%%%%%%%%%%%%%%% 
%In time versus episodes graph (refer Figure \ref{Map: Bridge-26x18 Against AI-Random} and \ref{Map: Bridge-26x18 Against AI-Follower}) the plot for Darmok2 is not showing much variation with respect to time for every episode because Darmok2 has knowledge of human traces and action plan and it can quickly win the match. When compared to Darmok2, SARSA takes more time to win because initially it has no knowledge about the game. But after some episodes the winning ratio for the SARSA increases. Q-Learning takes more time to win because it explores the states and follow the best path always not optimal or safe path. In RTS games there is no fixed best path because each time the states are different in different episodes. SARSA performs well because it checks the current state with the best action retrieved. If it will take to the safe state then only it chooses that retrieved action.
%%%%%%%%%%%%%%%%%%%%%%%%%%%%%%%%%

\begin{figure}
\centering
\includegraphics[width=2.5in, height=1.3in]{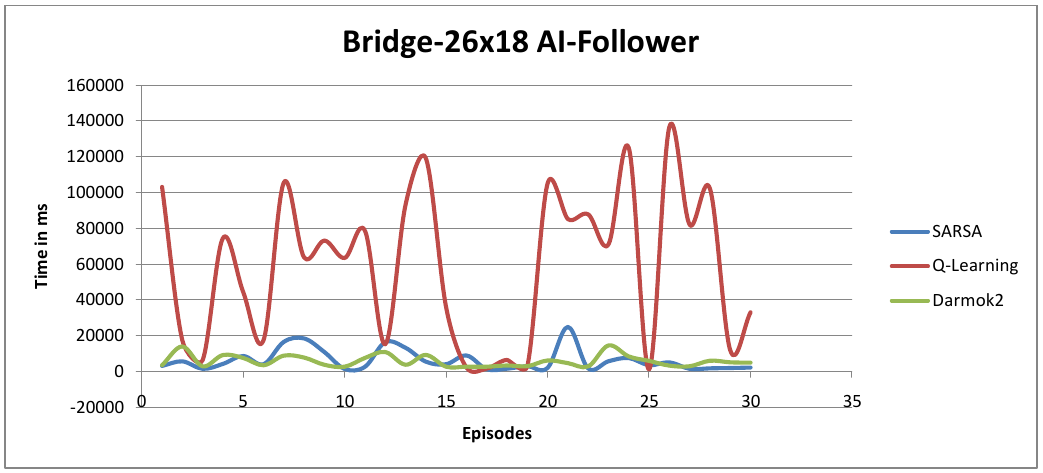}\\
\includegraphics[width=2.5in, height=1.3in]{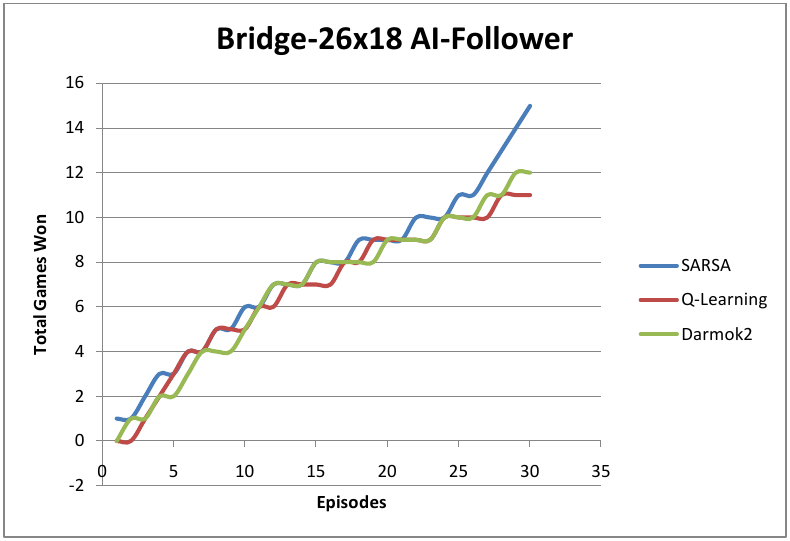} 
\caption{Map: Bridge-26x18 Against AI-Follower} \label{Map: Bridge-26x18 Against AI-Follower}
\end{figure}
\begin{figure}
\centering
\includegraphics[width=2.5in, height=1.3in]{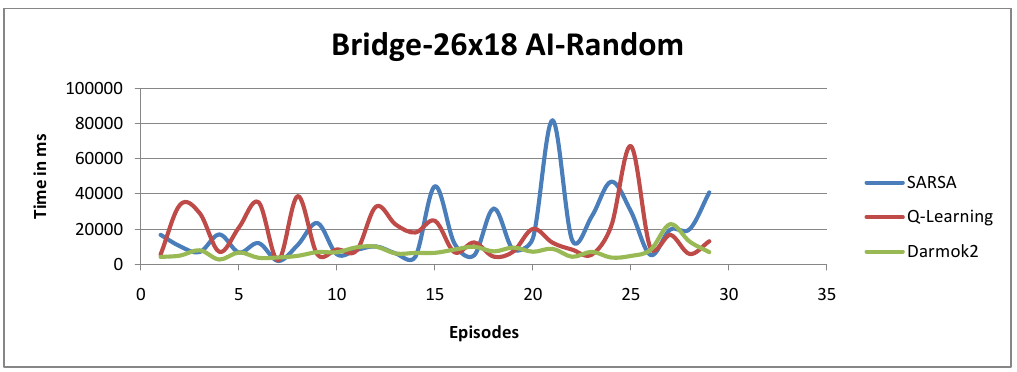}\\
\includegraphics[width=2.5in, height=1.3in]{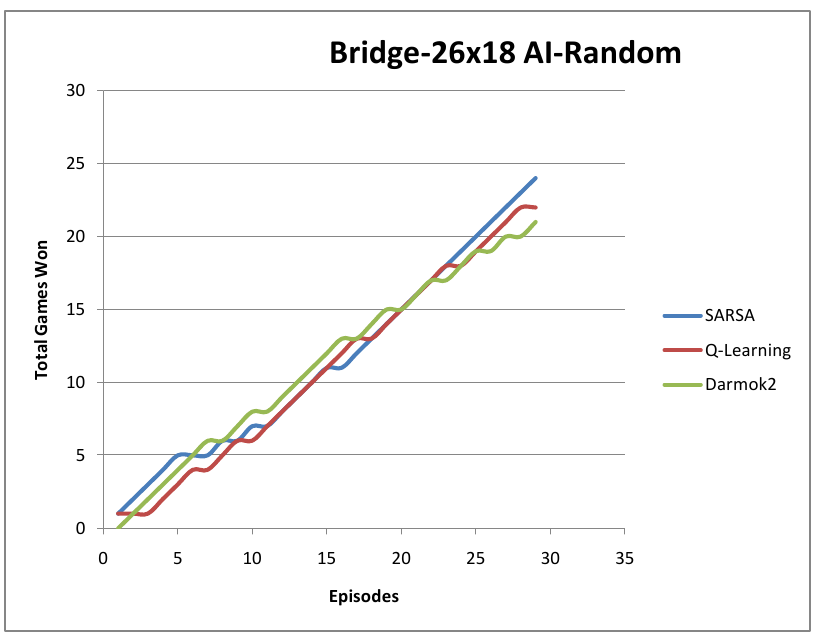} 
\caption{Map: Bridge-26x18 Against AI-Random} \label{Map: Bridge-26x18 Against AI-Random}
\end{figure}
\begin{figure}
\centering
\includegraphics[width=2.5in, height=1.3in]{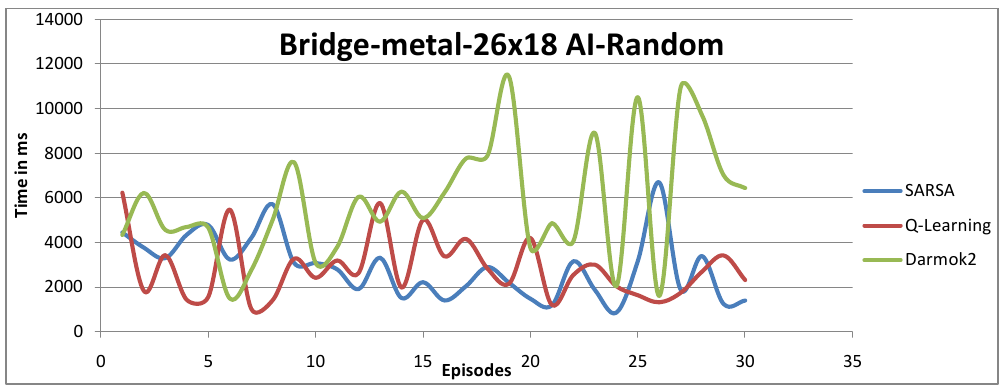}\\
\includegraphics[width=2.5in, height=1.3in]{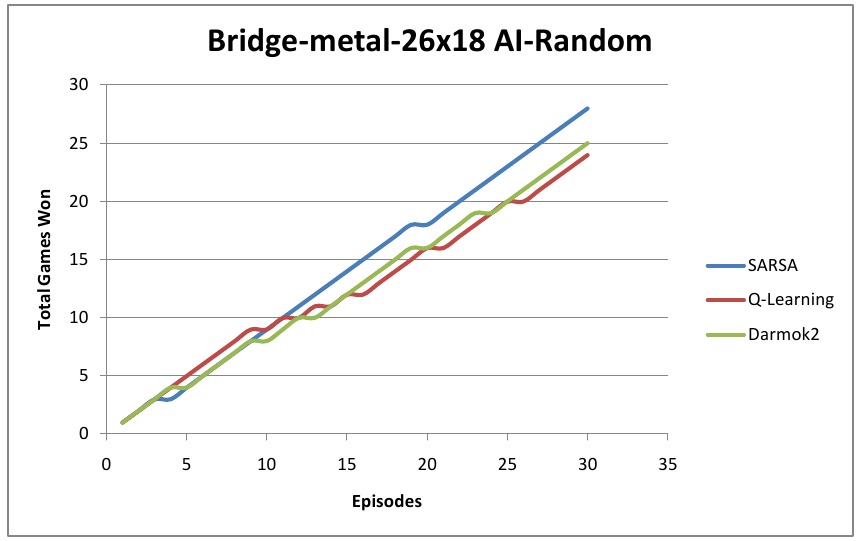} 
\caption{Map: Bridge-metal-26x18 Against AI-Random} \label{Map: Bridge-metal-26x18 Against AI-Random}
\end{figure}
\begin{figure}
\centering
\includegraphics[width=2.5in, height=1.3in]{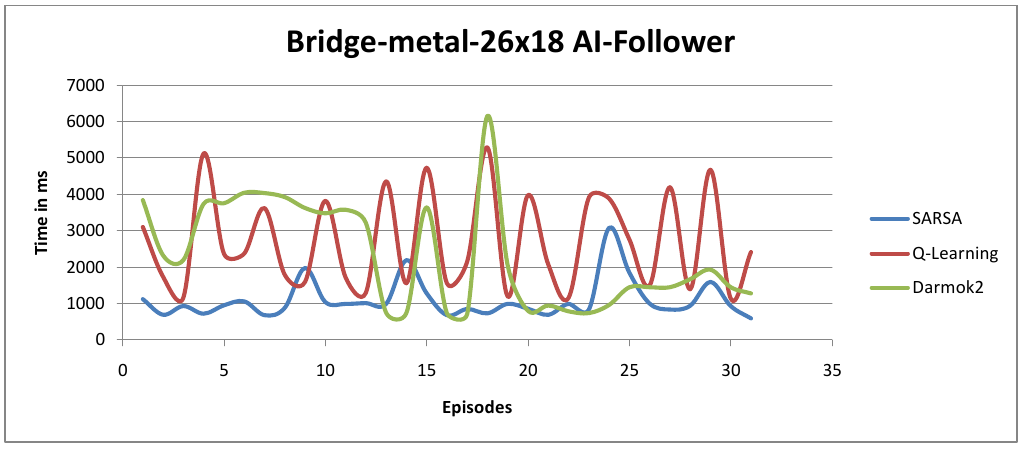}
\includegraphics[width=2.5in, height=1.3in]{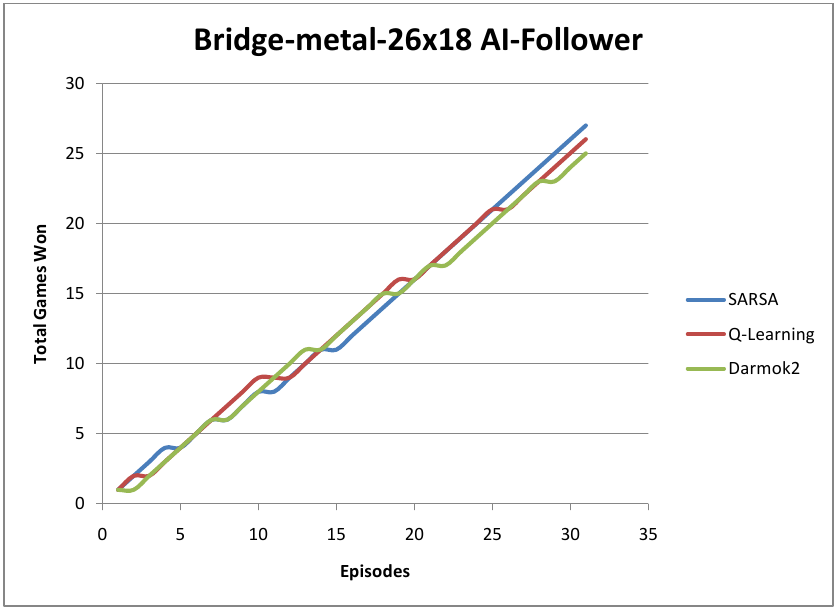} 
\caption{Map: Bridge-metal-26x18 Against AI-Follower} \label{Map: Bridge-metal-26x18 Against AI-Follower}
\end{figure}

\subsubsection{Map: Bridges-34x24}
\begin{figure}[h]
\centering \includegraphics[width=2.5in, height = 1.3in]{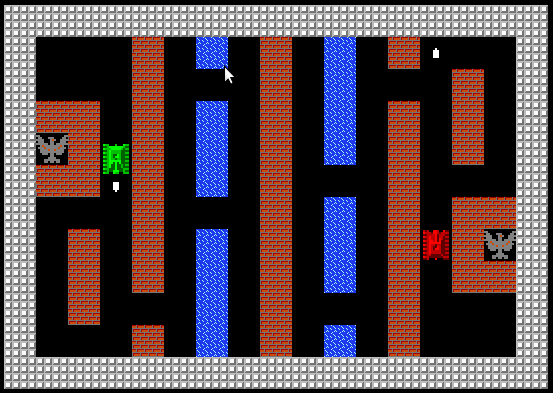}
\caption{Map:Bridge-Metal-34x24} \label{Map:Bridge-Metal-34x24}
\end{figure}

This is the most complex map (refer Figure \ref{Map:Bridge-Metal-34x24}) among all on which we have performed our evaluation because of its size and the structure. It is a 34x24 map and it has $34^{2}$x$24^{2}$ search spaces. It contains many brick wall and water bodies. Brick wall can be destroyed by firing. Its size and water bodies makes it a difficult and complex map.

\begin{figure}[h]
\centering
\includegraphics[width=2.5in, height=1.2in]{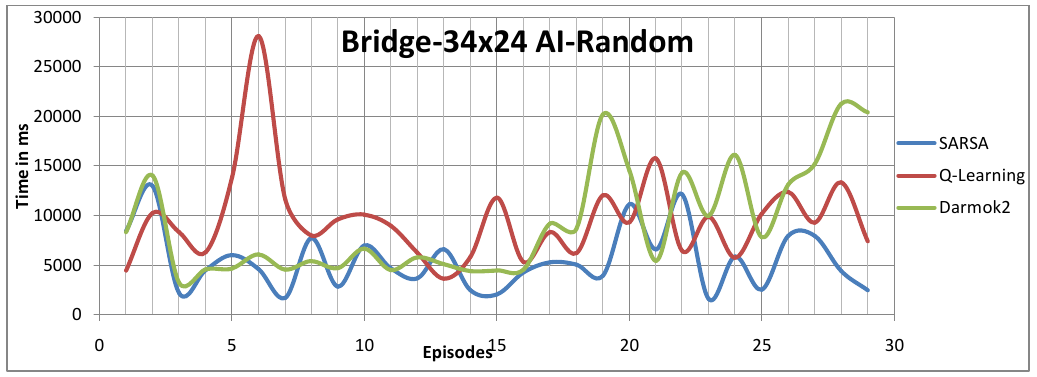}\\
\includegraphics[width=2.5in, height=1.2in]{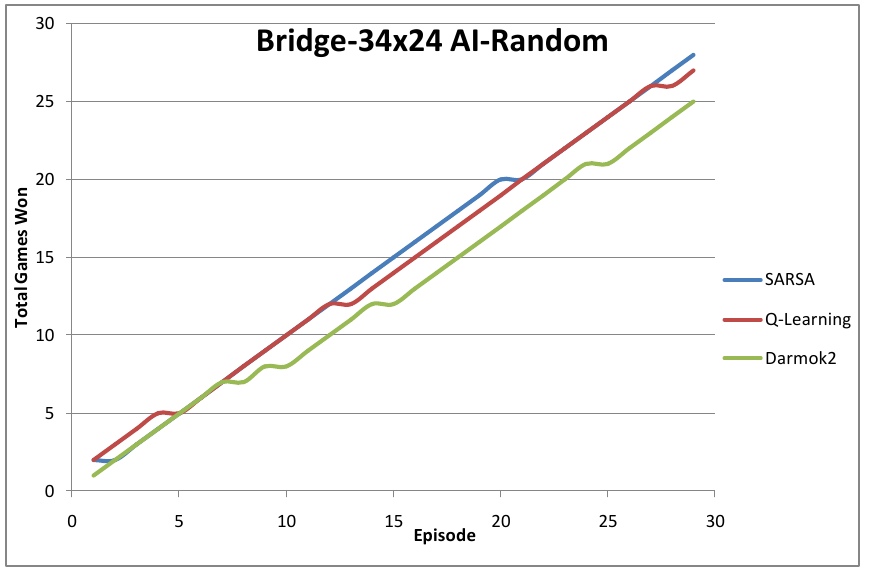}
\caption{Map: Bridges-34x24 Against AI-Random} \label{Map: Bridges-34x24 Against AI-Random}
\end{figure}

\begin{figure}[h]
\centering
\includegraphics[width=2.5in, height=1.2in]{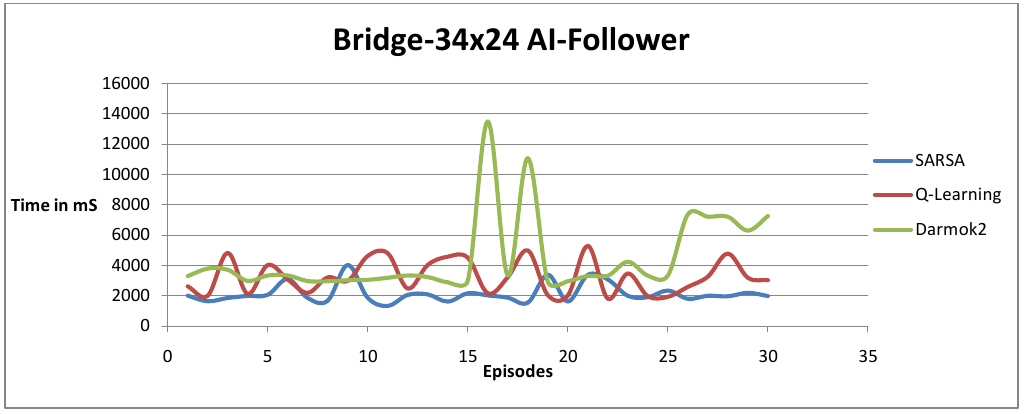}\\
\includegraphics[width=2.5in, height=1.2in]{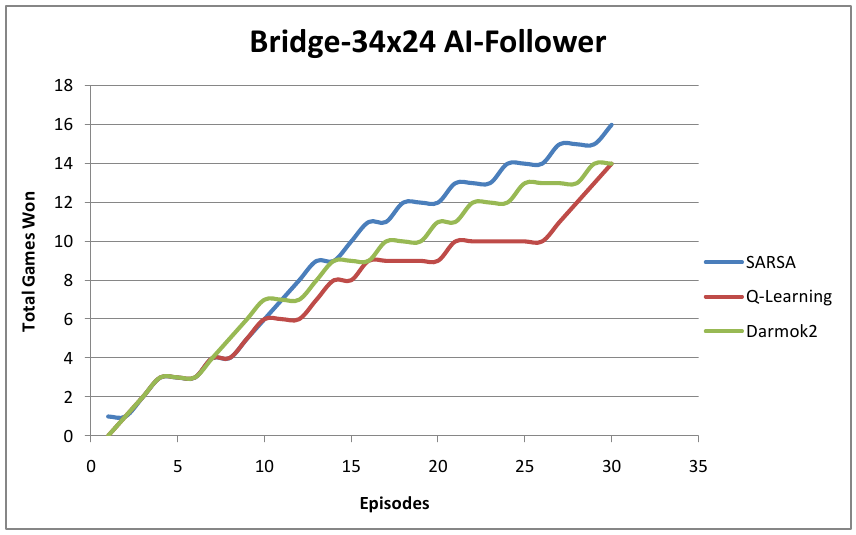}
\caption{Map: Bridges-34x24 Against AI-Follower} \label{Map: Bridges-34x24 Against AI-Follower}
\end{figure}

In time versus episodes graph (refer Figure \ref{Map: Bridges-34x24 Against AI-Random} and \ref{Map: Bridges-34x24 Against AI-Follower}) the plot (refer Figure \ref{Map: Bridge-26x18 Against AI-Random} and \ref{Map: Bridge-26x18 Against AI-Follower}) is showing that time to win the game for all strategies varies for every episodes. This map has more water bodies so it is difficult to learn a strategy to win quickly. Against AI-random the performance of all the strategies are close while in case of AI-follower SARSA performs well and wins more game in compared to Q-learning and Darmok2.  

\subsection{S3}
\begin{figure}[h]
\centering
\vspace{0.2cm}
\framebox{\includegraphics[width=6.5cm]{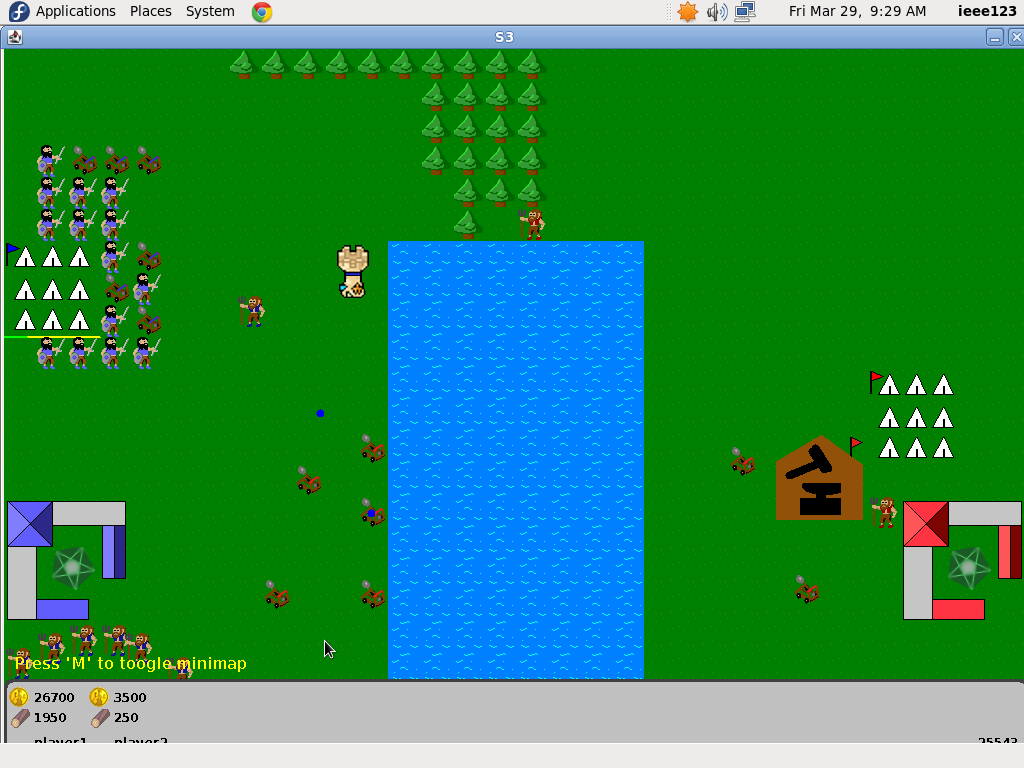}}
\caption{Snapshot of an S3 Game Map:GOW}
\label{GOW}
\end{figure}

The maps related to \textit{S3} are more complex than that of Battlecity. We evaluated our approach on various maps against several built-in AI player. In our experiments we built RL agent for S3 game using relative reward function with the Q-learning and SARSA approach as discussed earlier. RL-agent learn by playing 10 games against built-in-ai called \emph{ai-catapult-rush} for the simple map NWTR1 (refer Figure \ref{GOW}) using two approaches Q-Learning and SARSA. The state-action pair values (Q-Values) are updated while playing (or Learning as discussed earlier RL-Agent also learns while playing). Using this updated Q-Values RL-Agent plays games against \emph{ai-catapult-rush} as well as another type of built-in-ai called \emph{ai-rush}. 
\begin{itemize}
\item \emph{ai-catapult-rush} is the built-in-ai that builds barracks and lumber-mills at the starting, this has two peasants for harvesting gold, and two for harvesting wood. Then it starts building catapults nonstop and also attacks after a while. After sometime it increases the number of peasants to 3, and starts building the second barrack. It also looks for goldmines where there gold is still available. Also, it sends catapults to attack enemies.  
\item \emph{ai-rush} is the built-in-ai that builds a barrack at the starting. There are two peasants at the starting  for harvesting gold and wood. After building the barrack ai-rush trains the footmen. When there are two trained footmen it starts attacking.
\end{itemize}

For our experiment we used three type of maps (refer Figure \ref{snapshot of BattleCity and 2D map}, \ref{BattleCity_S3} and \ref{GOW}) according to difficulty level (easy-NWTR2, medium-NWTR6 and difficult-GOW). We performed our experiments with five games against two built-in-ai wherein the two approaches are Q-Learning and SARSA for each map. The comparison statistics are given in Table \ref{comparison}. We observed that RL-agent with SARSA wins most of the games. Q-Learning and the previous approach (Darmok2)~\cite{Santiago} performs almost the same but not better than SARSA. For S3 also SARSA gives the best results. 

Table \ref{comparison} shows the results comparison.
\begin{table*}[ht]
\small
\centering
\caption{ Comparison of SARSA and Q-Learning with Darmok2 }
\label{comparison}
\begin{tabular}{|c|c||c|c|c|c|c|}
\hline
map & Approach &Epoch 1 & Epoch 2 & Epoch 3 &Epoch 4 &Epoch 5\\
\cline{3-7}
&& \multicolumn{5}{|c|}{\textbf{against ai-catapult}} \\
\hline 
        NWTR2  & SARSA & won  &won &won &won &won \\ 
        NWTR2  & Q-Learning & lost &won &won &draw &won \\ 
        NWTR2  & Darmok2 & won &draw &won &won &lost \\ 
        NWTR6  & SARSA  & lost  &draw &won &won &won \\ 
        NWTR6  & Q-Learning & won   &lost &draw &lost &won \\ 
        NWTR6  &Darmok2 & won &lost &won &lost &won \\ 
        GOW & SARSA & draw &lost &won &draw &won \\ 
        GOW & Q-Learning & lost &lost &won &lost &won \\ 
        GOW & Darmok2 & won &lost &won &lost &draw \\ 
\hline
&&\multicolumn{5}{|c|}{\textbf{against ai-rush }}\\
\hline
        NWTR2  & SARSA    & won    &won &won &won &won      \\ 
        NWTR2  & Q-Learning    & won    &draw &won &won &won     \\ 
        NWTR2  & Darmok2     & won    &won &won &lost &won     \\ 
        NWTR6  & SARSA    & won &draw &won &won &won     \\ 
        NWTR6  & Q-Learning   & lost  &lost &won &won &won     \\ 
        NWTR6  & Darmok2     & won  &draw &won &lost &won     \\ 
        GOW  & SARSA     & draw   &won &won &won &won   \\ 
        GOW &   Q-Learning   & lost &won &draw &won &won     \\ 
        GOW  & Darmok2     & won &lost &won &lost &won    \\ 
\hline
\end{tabular}
\end{table*}
By analyzing  the results shown in Table \ref{comparison} we can see that in most of the maps SARSA has won or drawn the game. The maps where it has lost we found that the built-in-ai was a quick attacker and RL-agent was not able to produce enough number of troops to defend while the enemy was attacking. The RL agent was basically trying to find a way to enter through the wall of trees. 
In some maps we have shown the results as drawn. This means that resources like wood and gold of both player and enemy got finished and only peasants were left out at both the sides and they cannot do anything without the gold and wood.

When compared to previous research on Darmok2~\cite{Santiago}, where pre-prepared strategies are used to play the game and plan adaption module is used to switch strategies in this research RL-Agent quickly switches the strategies while playing, even though we used a simple map for training the RL-Agent.

\section{Conclusions}
\label{section:conclusions}

In this paper we proposed a reinforcement learning model for real-time strategy games. In order to achieve this end we make use of two reinforcement learning algorithms SARSA and Q-Learning. %To the best of our knowledge reinforcement learning (RL) has not been applied for real-time strategy games though there are works relating RL and turn-based games. 
The idea is to get the best action using one of the RL algorithms so as to not make use of the \emph{traces} generated by the players. In previous works on real-time strategy games using "on line case based learning" human traces form an important component in the learning process. In the proposed method we are not making use of any previous knowledge like traces and therefore we follow an unsupervised approach.
This research is with regard to getting the best action using two algorithms (SARSA and Q-Learning) which comes under Reinforcement Learning without the traces generated by the player as proposed in the previous work "on line case based learning" using Darmok2. 
Another major contribution of  our work is the reward function. Rewards are calculated by two types of reward functions called conditional and generalized reward function. The sensor information related to game is used for calculating the rewards. The reward values are further used by the two RL algorithms

SARSA and Q-Learning. These algorithms make policies according to the reward for the state-action pair. RL agent choose the action using these policies. We evaluated our approach successfully in two different game domains (BattleCity and S3) and observed that reinforcement learning performs better than previous approaches in terms of learning time and winning ratio. In particular SARSA algorithm takes lesser time to learn and start winning very quickly than Q-Learning and that too for complex maps.
%Our future work is to evaluate our approach in domains that are more complex than S3. %Additionally, identifying the areas which can combine with our approach to improve the performance with vast search space, so that we can process search space efficiently and fast to improve the performance of reinforcement learning.

\noindent{\includegraphics[width=1in,height=1.7in,clip,keepaspectratio]{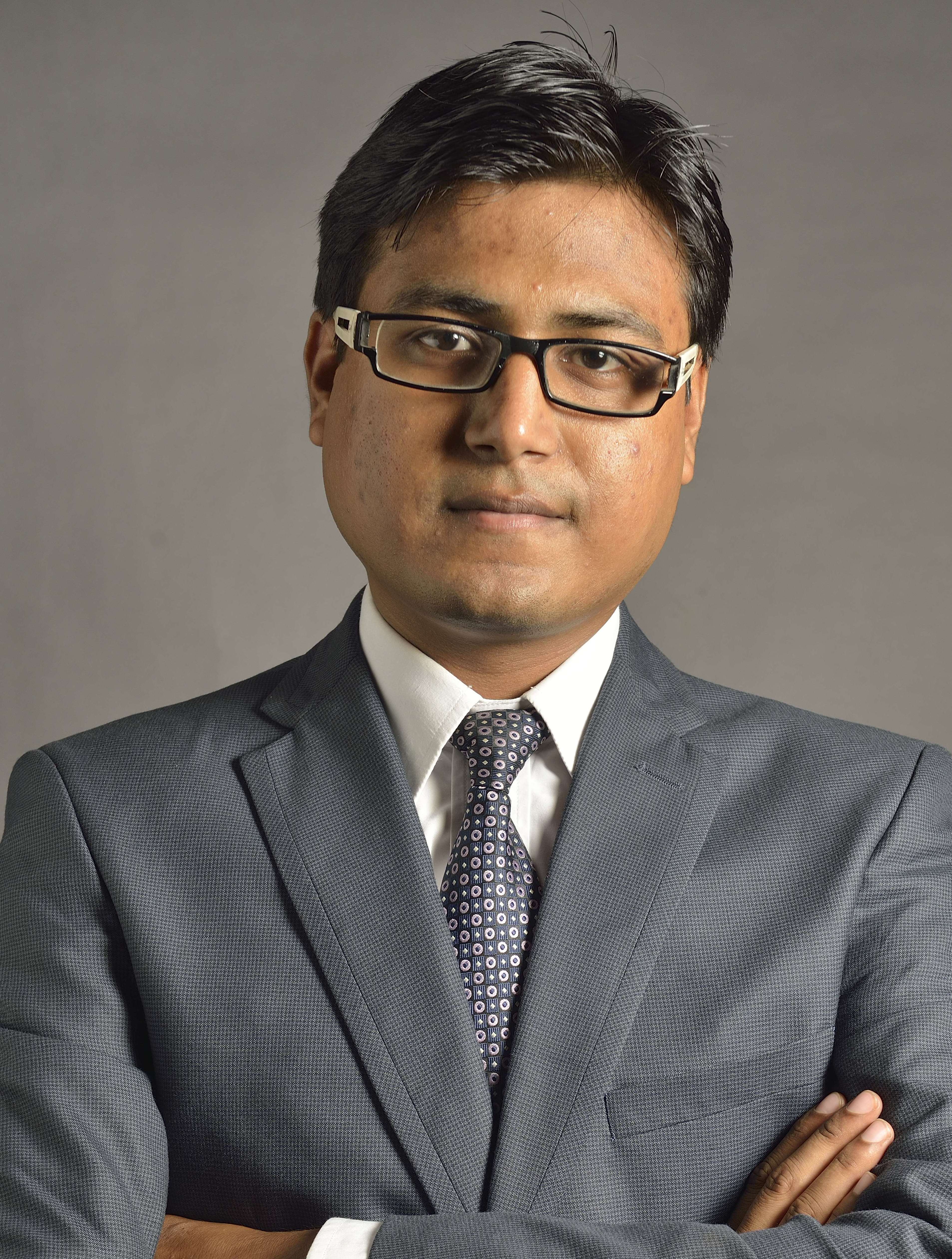}}
\begin{minipage}[b][1in][c]{1.8in}
{\centering{\bf{Harshit Sethy }}is the Co-founder and Chief Technology Officer of Gymtrekker Fitness Private Limited, Mumbai, India. He received his Masters degree in Artificial Intelligence from University of Hyderabad.}\\\\
\end{minipage} 
\\\\
%---------------------------------------------------------------------------------------------------------------------
%\begin{biography}
\noindent{\includegraphics[width=1in,height=1.7in,clip,keepaspectratio]{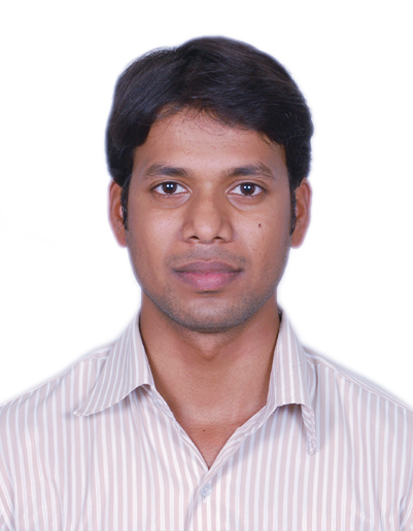}}
\begin{minipage}[b][1in][c]{1.8in}
{\centering{\bf {Amit Patel}} is currently the Assistant Professor, Rajiv Gandhi University of Knowledge Technologies, IIIT Nuzvid, Krishna. He obtained his Bachelor of Technology from Uttar Pradesh Technical University. He received his}\\\\
\end{minipage}
Masters degree in Artificial Intelligence from University of Hyderabad, Hyderabad.


\begin{thebibliography}{1}

\bibitem{RSSuttan}
R. S. Sutton and A. G. Barto,{\it Reinforcement Learning: An Introduction}.
A book publisher MIT Press; 1998.

\bibitem{Genter}
Katie~Long Genter. Using first order inductive learning as an alternative to a simulator in a game arficial intelligence. under-graduate thesis. In {\it Georgia Institute of Technology}, pages 1--2, May 2009.

\bibitem{AshwinRam}
Katie~Long Genter, Santiago Onta{\~n}{\'o}n, and Ashwin Ram. Learning opponent strategies through first order induction. In : {\it FLAIRS Conference}, pages 1--2, 2011.

\bibitem{Gomez}
P.~P. Gomez-Martin, D.~Llanso, M.~A. Gomez-Martin, Santiago Onta{\~n}{\'o}n,
  and Ashwin Ram. Mmpm: a generic platform for case-based planning research.
In : {\it ICCBR 2010 Workshop on Case-Based Reasoning for Computer Games}, pages 45-54, July 2010.

\bibitem{Stefan}
Stefan Wender and Ian Watson. Using Reinforcement Learning for City Site Selection in the Turn-Based Strategy Game Civilization IV. In : {\it Computational Intelligence and Games (CIG-2008)}, pages 372-377, 2008.

\bibitem{Janet}
Janet~L. Kolodner. An introduction to case-based reasoning. In : {\it Artificial Intelligence Review}, pages 3-34, 1992.

\bibitem{Santi}
Santi Onta{\~n}{\'o}n, Kinshuk Mishra, Neha Sugandh, and Ashwin Ram. On-line case-based planning. In : {\it Computational Intelligence}, pages 84-119, 2010.

\bibitem{Onta}
Santiago Onta{\~n}{\'o}n, K.Bonnette, P.Mahindrakar, M.~A. Gomez-Martin,
  Katie~Long Genter, J.Radhakrishnan, R.Shah, and Ashwin Ram. Learning from human demonstrations for real-time case-based planning. In : {\it STRUCK-09 Workshop, colocated with IJCAI}, pages 2-3, 2011.

\bibitem{Neha}
Neha Sugandh, Santiago Onta{\~n}{\'o}n, and Ashwin Ram. On-line case-based plan adaptation for real-time strategy games. In :{\it Association for the Advancement of Artificial Intelligence (AAAI-2008)}, pages 1-2. AAAI Press, 2008.

\bibitem{Santiago}
Santiago~Onta{\~n}{\'o}n Villar. D2 documentation. pages 1-6, May 2010, http://heanet.dl.sourceforge.net/project/dar-mok2/D2Documentation.pdf.

\bibitem{Ponsen}
Marc Ponsen and Pieter Spronck
\newblock Improving Adaptive Game AI with Evolutionary Learning
\newblock In {\em Computer Games: Artificial Intelligence, Design and Education}, pages 389-396, 2004.

\bibitem{Marthi}
Bhaskara Marthi, Stuart Russell, David Latham and Carlos Guestrin Concurrent hierarchical reinforcement learning
Turn-Based Strategy Game Civilization IV. In {\em International Joint Conference on Artificial Intelligence, Edinburgh, Scotland}, pages 1652-1653, 2005.

\bibitem{Pranay}
Pranay M. Game AI : Simulator Vs Learner in Darmok2. In {\it University of Hyderabad as M.Tech. thesis} , 2013.

\end{thebibliography}
\end{document}